\documentclass[11pt,a4paper]{article}
\usepackage[hyperref]{acl2021}
\usepackage{times}
\usepackage{latexsym}

\usepackage{multirow}
\usepackage{subcaption}
\usepackage{adjustbox}
\usepackage{makecell}
\usepackage{amssymb}% http://ctan.org/pkg/amssymb
\usepackage{pifont}% http://ctan.org/pkg/pifont
\newcommand{\cmark}{\ding{51}}%
\newcommand{\xmark}{\ding{55}}%
\usepackage{microtype}

\aclfinalcopy % Uncomment this line for the final submission
\title{Continual Mixed-Language Pre-Training for Extremely Low-Resource Neural Machine Translation}

\author{Zihan Liu, Genta Indra Winata, Pascale Fung \\
Center for Artificial Intelligence Research (CAiRE)\\
Department of Electronic and Computer Engineering\\
The Hong Kong University of Science and Technology, Clear Water Bay, Hong Kong\\
\texttt{zihan.liu@connect.ust.hk, pascale@ece.ust.hk}}

\date{}

\begin{document}
\maketitle
\begin{abstract}
% motivation
% low-resource issue

The data scarcity in low-resource languages has become a bottleneck to building robust neural machine translation systems. Fine-tuning a multilingual pre-trained model (e.g., mBART~\cite{liu2020multilingual}) on the translation task is a good approach for low-resource languages; however, its performance will be greatly limited when there are unseen languages in the translation pairs. In this paper, we present a continual pre-training (CPT) framework on mBART to effectively adapt it to unseen languages. We first construct noisy mixed-language text from the monolingual corpus of the target language in the translation pair to cover both the source and target languages, and then, we continue pre-training mBART to reconstruct the original monolingual text. Results show that our method can consistently improve the fine-tuning performance upon the mBART baseline, as well as other strong baselines, across all tested low-resource translation pairs containing unseen languages. Furthermore, our approach also boosts the performance on translation pairs where both languages are seen in the original mBART's pre-training. The code is available at \url{https://github.com/zliucr/cpt-nmt}.

\end{abstract}

\section{Introduction}
% 1. Motivation of this work
% 1.1 Low-resource issue in NMT and pre-trained models
% 1.2 Unseen language issue and So many languages in this world and multilingual pre-trained model is impossible to include all of them
% 1.3 pre-training from scratch is time-consuming
% 1.4 we want to propose a method to leverage the off-the-shelf pre-trained model and generalize it to all low-resource languages
Neural machine translation (NMT)~\cite{bahdanau2015neural,luong2015effective,vaswani2017attention} has a poor generalization ability to low-resource languages where large monolingual and parallel corpora are not available. 
Recently, leveraging multilingual pre-trained models~\cite{song2019mass,liu2020multilingual,lin2020pre} as the starting checkpoints has shown to be effective at building low-resource NMT systems. However, the effectiveness of the pre-training will be vastly limited for low-resource languages that are not in the list of pre-training languages. Given the fact that there are more than 7000 languages around the world~\cite{austin2011cambridge}, it is almost impossible for a multilingual model to include all languages. And it is expensive and time-consuming to pre-train another model from scratch so as to include the languages we need. 
To address this issue, we propose to leverage the advantages of an off-the-shelf multilingual pre-trained model and focus on better generalizing it to any low-resource language pair. In this paper, we use mBART~\cite{liu2020multilingual} as the multilingual pre-trained model, given its effectiveness at building low-resource NMT systems.

\begin{figure*}[!ht]
  \centering
  \includegraphics[width=0.999\linewidth]{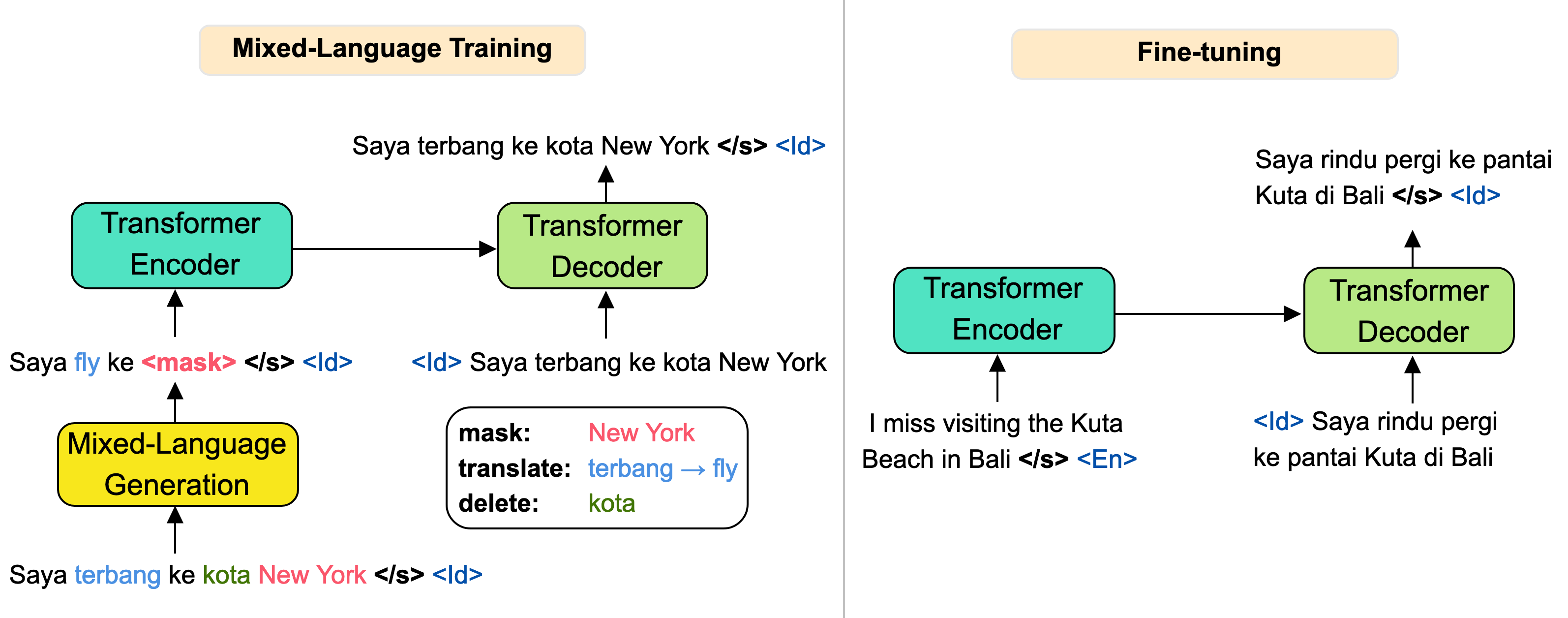}
%   \caption{Framework for Continual Pre-Training: Original mBART pre-training (left), our mixed-language pre-training (center), and fine-tuning on the translation task (right).}
  \caption{An illustration of adapting mBART to the En-Id translation pair: Continual pre-training with mixed-language training (left) and fine-tuning on the translation task (right).}
  \label{fig:training}
\end{figure*}

% 2. Description of our methods
To simulate the problem, we suppose that we need an NMT system on a low-resource translation pair, and at least one of the languages in the translation pair is an unseen language for the pre-trained model.
% To simulate the problem, we train an NMT system on the extremely low-resource setting by fine-tuning a pre-trained model, where at least one of the languages is unseen on the pre-trained model. 
To adapt mBART into unseen languages in the NMT task, we propose to conduct a continual pre-training (\textbf{CPT}) on it with mixed-language training (\textbf{MLT}).
% The general idea is to conduct a Continual Pre-Training (\textbf{CPT}) on a multilingual pre-trained model (mBART) and adapt it to the unseen low-resource languages.
% Concretely, we utilize a bilingual dictionary to generate mixed-language sentences and simultaneously corrupt the text by deleting some tokens to introduce noise. Then, we take mixed-language sentences as the input of our CPT, and we reconstruct the original monolingual text by following the mBART's objective function (i.e., denoising auto-encoding).
Concretely, we first follow the noise function used in~\citet{liu2020multilingual} to corrupt the monolingual text of the target language in the translation. Then, we utilize a bilingual dictionary to generate mixed-language sentences and simultaneously delete some tokens based on the corrupted text. After that, we conduct the CPT on mBART to reconstruct the original monolingual text. After the CPT, we follow~\citet{liu2020multilingual} to directly fine-tune mBART on the parallel data of the translation pair.
The purpose of producing mixed-language sentences is to make a rough alignment between the languages in the translation pair. Conducting the token deletion is to increase the difficulty of the reconstruction task and the diversity of the noisy mixed-language text, which force the model to quickly learn an unseen language. 
% The alignments facilitate the cross-lingual transfer on the model.

% 3. Our settings and results
We consider an extremely low-resource setting where we have very few parallel data (10k) for low-resource translation pairs and very few monolingual data (100k) for each language in the translation. Experimental results show that our proposed pre-training approach is able to consistently outperform the mBART baseline as well as other pre-training baselines across all tested translation pairs that contain unseen languages. Interestingly, we observe that the continual mixed-language pre-training is even beneficial for a translation pair where both languages are in the mBART's pre-training list.
Results also show that mBART can achieve better zero-shot performance after applying the CPT with MLT, which illustrates that the mixed-language pre-training is able to make a better alignment. Furthermore, we investigate our method in terms of various low-resource settings where different amounts of parallel and monolingual data are available, and experimental results show that the effectiveness of our approach can be further improved when a larger pre-training corpus is available.
% conduct experiments with different amounts of monolingual data for pre-training as well as parallel data for fine-tuning, and we show that the improvements made by the CPT can be consistently increased when a larger pre-training corpus is available.

% 4. Contributions
% The main contributions of this paper are summarized as follows:
The contributions of this paper are summarized as follows:
\begin{itemize}
    \item To the best of our knowledge, we are the first to investigate how to effectively adapt a multilingual pre-trained model to unseen languages for the NMT task.
    \item We show that our proposed method can consistently surpass strong baselines across all the tested translation pairs.
    \item We conduct in-depth experiments and analyses in terms of different low-resource settings and the effectiveness on the various components of our method.
\end{itemize}

\section{Methodology}
In this section, we first give a brief overview of the mBART model~\cite{liu2020multilingual}, and then we introduce our proposed method that aims to adapt mBART to unseen languages in the translation task.

\subsection{Model: mBART} \label{mbart_model}
The mBART model follows the sequence-to-sequence (Seq2Seq) pre-training scheme of the BART model~\cite{lewis2020bart} (i.e., reconstructing the corrupted text) and is pre-trained on large-scale monolingual corpora in 25 languages.
Two types of noises are used to produce the corrected text. The first is to remove text spans and replace them with a mask token, and the second is to permute the order of sentences within each instance.

Thanks to the large-scale pre-training on multiple diverse languages, the mBART model has shown its strength at building low-resource NMT systems by being fine-tuned to the target language pair, and it is also shown to possess a powerful generalization ability to languages that do not appear in the pre-training corpora~\cite{liu2020multilingual}.

\subsection{Continual Pre-Training}
Despite the powerful adaptation ability that mBART possesses, we argue that its performance on unseen languages is still sub-optimal since it has to learn these languages from scratch.
Therefore, we propose to conduct the continual pre-training (CPT) on the mBART model to improve its adaptation ability to unseen languages. 
The process of this additional pre-training task is illustrated in Figure~\ref{fig:training}, and the details are described as follows.

\paragraph{Pre-Training}
We denote \texttt{lang$_1$$\rightarrow$lang$_2$} as the needed translation pair, where \texttt{lang$_1$} is the source language and \texttt{lang$_2$} is the target language, and at least one of them is an unseen language for the mBART model. 
The CPT can be considered as maximizing $L_\theta$:
\begin{equation}
    L_\theta = \sum_{X \in D_2} \log P(X|f(X); \theta), \label{eq1}
\end{equation}
where $\theta$ is initialized with mBART's parameters, $D_2$ denotes a collection of monolingual documents in \texttt{lang$_2$}, and $f$ is a function to generate noisy mixed-language text that contains both \texttt{lang$_1$} and \texttt{lang$_2$}.

\paragraph{Noisy Mixed-Language Function ($f$)} \label{noisy_cs_function}
Given a monolingual instance $X$, we first use the noise function (denoted as $g$, described in §\ref{mbart_model}) used in~\citet{liu2020multilingual} to corrupt the text, and then we use a dictionary of \texttt{lang$_2$} to \texttt{lang$_1$} to assist in the function of producing mixed-language sentences (denoted as $h$). Specifically, after the processing of the noise function $g$, if the non-masked tokens in \texttt{lang$_2$} exist in the dictionary, we set a probability to replace it with its translation in \texttt{lang$_1$}. If it is not being replaced, there is a 50\% chance that we will directly delete this token, and otherwise, we keep the original token in \texttt{lang$_2$}. More formally, function $f$ (in Eq.~(\ref{eq1})) can be considered as the combination of two functions:
\begin{equation}
    f(X) = h(g(X)).
\end{equation}
Notice that \texttt{lang$_2$} is not always the unseen language (i.e., \texttt{lang$_1$} could be the only unseen language). Since the inputs are mixed with the tokens in \texttt{lang$_1$} and \texttt{lang$_2$}, the model can always learn the unseen language.

The reason why we choose to reconstruct \texttt{lang$_2$} instead of \texttt{lang$_1$} is because \texttt{lang$_2$} is the target language that the decoder needs to generate in the translation task, and reconstructing \texttt{lang$_2$} in the pre-training makes the model easier to adapt to the \texttt{lang$_1$$\rightarrow$lang$_2$} translation pair.
We leverage the noise function $g$ since it has shown its effectiveness at helping pre-trained models to obtain language understanding ability. The intuition of producing mixed-language text for inputs is to roughly align \texttt{lang$_1$} and \texttt{lang$_2$}, since the model needs to understand the tokens of \texttt{lang$_1$} so as to reconstruct the translations in \texttt{lang$_2$}. 
The purpose of not replacing all tokens in the dictionary with their translations is to increase the variety of the mixed-language text, and given that there will be plenty of frequent words (e.g., stopwords), replacing all of them with the corresponding translations could make the sentences unnatural, and the translations of the frequent words in \texttt{lang$_1$} would likely not match the context in \texttt{lang$_2$}. In addition, adding a probability to delete the original token in function $h$ is to inject extra noise and further increase the diversity of the generated mixed-language text.

\section{Experimental Settings}
\subsection{Datasets}
% bilingual
We conduct experiments on 12 low-resource language pairs from OpenSubtitles~\cite{lison2016opensubtitles2016}, resulting in 24 directed translation pairs in total.
% Each pair has at least one language that is an unseen language for mBART. 
Each pair has an unseen language for mBART.
Concretely, there are 12 translation pairs (out of 24) containing English and another unseen language (Indonesian (Id), Ukrainian (Uk), Bengali (Bn), Afrikaans (Af), Tamil (Ta), Thai (Th) $\leftrightarrow$ English (En)), and the rest of the 12 pairs contain two unseen languages (Id $\leftrightarrow$ Ta, Bn $\leftrightarrow$ Th, Bulgarian (Bg) $\leftrightarrow$ Ta, Id $\leftrightarrow$ Bn, Macedonian (Mk) $\leftrightarrow$ Th, and Slovak (Sk) $\leftrightarrow$ Swedish (Sv)). 
In addition, we evaluate the translation pairs (En $\leftrightarrow$ Gujarati (Gu) and En $\leftrightarrow$ Kazakh (Kk) (WMT19)), where both languages are in mBART's pre-training list.

% monolingual
To produce noisy mixed-language sentences, we collect monolingual corpora for the target languages from Wikipedia, and we utilize the bilingual dictionaries from MUSE~\cite{lample2018word}\footnote{https://github.com/facebookresearch/MUSE} for the En-X and X-En pairs. For a dictionary (denoted as X-Y) that is not available in MUSE (English is not in the pair in this case), we first obtain the token list of language X from the X-En dictionary in MUSE, and then construct the X-Y dictionary utilizing Google Translate\footnote{https://translate.google.com. The constructed dictionaries will be released in \url{https://github.com/zliucr/cpt-nmt}.} to translate the tokens from language X to Y.

\subsection{Low-Resource Settings}
We focus on an extremely low-resource setting, where we assume that only 10K parallel samples are available. 
Considering that obtaining a large size monolingual corpus could be difficult for some low-resource languages, we constrain the number of monolingual paragraphs to be as few as 100K (the size is $\sim$ 30MB). To do so, we randomly sample 10K parallel examples and 100K monolingual paragraphs from the available corpora.
In addition, we also conduct experiments with different numbers of parallel data (from 10K to 100K) and monolingual data (from 100K to 1M) to investigate the effectiveness of the proposed method in different levels of low-resource setting. As for the translation pairs En $\leftrightarrow$ Gu and En $\leftrightarrow$ Kk, we follow the settings in~\citet{liu2020multilingual} and use parallel data with a size of 10K and 91K for the En $\leftrightarrow$ Gu and En $\leftrightarrow$ Kk, respectively.

\begin{table*}[!t]
\renewcommand{\arraystretch}{1.19}
\centering
\begin{adjustbox}{width={0.999\textwidth},totalheight={\textheight},keepaspectratio}
\begin{tabular}{c|cccccccccccc}
\Xhline{2\arrayrulewidth}
\textbf{Language Pairs}     & \multicolumn{2}{c}{\textbf{En-Id}}        & \multicolumn{2}{c}{\textbf{En-Uk}}        & \multicolumn{2}{c}{\textbf{En-Bn}}        & \multicolumn{2}{c}{\textbf{En-Af}}        & \multicolumn{2}{c}{\textbf{En-Ta}}        & \multicolumn{2}{c}{\textbf{En-Th}}        \\
\textbf{Direction}          & $\leftarrow$ & $\rightarrow$ & $\leftarrow$ & $\rightarrow$ & $\leftarrow$ & $\rightarrow$ & $\leftarrow$ & $\rightarrow$ & $\leftarrow$ & $\rightarrow$ & $\leftarrow$ & $\rightarrow$ \\ \hline
mT5                & 8.15          & 6.98             & 4.53          & 0.68             & 1.50          & 0.34             & 7.68          & 8.83             & 1.98          & 2.15             & 2.87          & 2.19             \\
mBART              & 8.87         & 7.38            & 4.85          & 0.89             & 1.37          & 0.65             & 8.24          & 10.02            & 4.07          & 2.70             & 3.12          & 2.41             \\ \hline
CPT w/ Ori (Src)  & 9.05          & 7.41             & 5.49          & 1.11             & 1.90          & 0.76             & 8.29          & 9.32             & 3.80          & 4.05             & 3.17          & 3.16             \\
CPT w/ Ori (Tgt)  & 8.78          & 7.77             & 5.75          & 1.31             & 2.03          & 0.92             & 8.31          & 9.71             & 3.46          & 4.26             & 3.08          & 3.57             \\ \hline
CPT w/ MLT (Src) & 10.44         & 8.40             & 5.22          & 1.45             & 2.21          & \textbf{1.43}             & 8.58          & 10.12            & 4.28          & 5.05             & 3.42          & \textbf{4.80}             \\
CPT w/ MLT (Tgt) & \textbf{11.16}         & \textbf{10.30}            & \textbf{6.50}          & \textbf{1.48}             & \textbf{2.73}          & 1.25             & \textbf{10.56}         & \textbf{11.62}            & \textbf{6.21}          & \textbf{5.20}             & \textbf{3.85}          & 4.54             \\ \hline \hline
\textbf{Language Pairs}     & \multicolumn{2}{c}{\textbf{Id-Ta}}        & \multicolumn{2}{c}{\textbf{Bn-Th}}        & \multicolumn{2}{c}{\textbf{Bg-Ta}}        & \multicolumn{2}{c}{\textbf{Id-Bn}}        & \multicolumn{2}{c}{\textbf{Mk-Th}}        & \multicolumn{2}{c}{\textbf{Sk-Sv}}        \\
\textbf{Direction}          & $\leftarrow$ & $\rightarrow$ & $\leftarrow$ & $\rightarrow$ & $\leftarrow$ & $\rightarrow$ & $\leftarrow$ & $\rightarrow$ & $\leftarrow$ & $\rightarrow$ & $\leftarrow$ & $\rightarrow$ \\ \hline
mT5                & 0.83          & 0.45             & 0.00          & 0.21             & 0.33          & 0.22             & 0.10          & 0.07             & 0.32          & 0.23             & 0.44          & 1.83             \\
mBART              & 1.21          & 0.98             & 0.00          & 0.00             & 0.52          & 0.26             & 0.00          & 0.00             & 0.29          & 0.41             & 0.38          & 1.76             \\ \hline
CPT w/ Ori (Src)  & 0.93          & 1.49             & 0.00          & 0.52             & 0.39          & 0.30             & 0.41          & 0.22             & 0.48          & 0.60             & 0.73          & 1.57             \\
CPT w/ Ori (Tgt)  & 1.24          & 1.24             & 0.00          & 0.64             & 0.41          & 0.61             & 0.33          & 0.30             & 0.51          & 0.67             & 0.78          & 2.09             \\ \hline
CPT w/ MLT (Src) & 1.39          & \textbf{1.90}             & 0.00          & 0.09             & 0.66          & 0.52             & \textbf{0.54}          & \textbf{0.31}             & \textbf{0.73}          & \textbf{1.21}             & 0.83          & 2.21             \\
CPT w/ MLT (Tgt) & \textbf{2.52}          & 1.75             & \textbf{0.20}          & \textbf{0.66}             & \textbf{0.95}          & \textbf{0.85}             & 0.36          & \textbf{0.31}             & 0.69          & 1.15             & \textbf{0.99}          & \textbf{2.55}             \\ \Xhline{2\arrayrulewidth}
\end{tabular}
\end{adjustbox}
\caption{Fine-tuning performance on the 10K parallel data for the 24 translation pairs. All CPT methods utilize a corpus with a size of 100K paragraphs. The upper 12 pairs contain one unseen language for mBART (the other seen language is English), and the bottom 12 pairs contain two unseen languages. The CPT using our proposed method consistently outperforms all baseline models.}
\label{tab:main}
\end{table*}

\subsection{Models \& Baselines}
\paragraph{mBART}
We directly fine-tune the mBART model on the parallel data of the translation pair. Note that it is already a strong baseline since mBART is shown to possess a good generalization ability to unseen languages~\cite{liu2020multilingual}.

\paragraph{CPT w/ Ori (Src)}
We follow the \textbf{original} objective function of mBART (only using the noise function $g$ in §\ref{noisy_cs_function} to corrupt the text) to continue pre-training it on the \textbf{source} language of the translation pair.~\footnote{For example, in the Id $\rightarrow$ Ta translation, the source language is Id and the target language is Ta.} Then we directly fine-tune it on the translation parallel data.

\paragraph{CPT w/ Ori (Tgt)}
This baseline is the same as the previous one except that we continue pre-training mBART on the \textbf{target} language of the translation pair.

\paragraph{CPT w/ MLT (Src)}
Different from CPT w/ Ori, we use the noisy mixed-language function ($f$) to create noisy mixed-language text. 
However, different from what we propose in Eq.~(\ref{eq1}), it reverses the pre-training direction (i.e., it corrupts the text in the \textbf{source} language instead of the target language).

\paragraph{CPT w/ MLT (Tgt)}
This is our proposed method described in §\ref{noisy_cs_function}. We use Tgt or Src to distinguish the target or source language (in the translation pair), respectively, that mBART needs to reconstruct in the CPT.

\paragraph{mT5}
Like mBART, mT5~\cite{xue2020mt5} is also a multilingual pre-trained model using a Seq2Seq pre-training. It is pre-trained in 101 languages covering all the languages in our experimental settings. Note that we use the mT5-base (600M parameters) which has a similar size as mBART (610M parameters) to ensure the fair comparison.

\subsection{Training Details}
Given that the sizes of the pre-training data and the parallel data are relatively small, we freeze the first 8 layers (out of 12) of the encoder and the first 8 layers (out of 12) of the decoder in the CPT, as well as the fine-tuning processes (applied for both mBART and mT5), to avoid the over-fitting issue. Note that we still keep the embeddings layer unfrozen since the model needs to learn the embeddings for unseen languages.
For CPT, we control the probability of whether to replace a token with its translation to ensure around 30\% of tokens are replaced. 
In the CPT stage, we train with a dropout rate of 0.1, a batch size of 100, and a learning rate of 3e-5 for 5 epochs. In the fine-tuning stage, we train with a dropout rate of 0.3, a batch size of 32, and 2500 warm-up steps with a maximum learning rate of 5e-5 for all directions. We use the Adam optimizer~\cite{kingma2015adam} for both the CPT and fine-tuning processes.
We set the maximum fine-tuning epochs as 20, and the final model is selected based on the performance on the validation dataset. 
The final results are reported in the case-sensitive tokenized BLEU~\cite{papineni2002bleu}. We notice that the tokenizer of mBART is the same as that of XLM-R~\cite{conneau2020unsupervised} which covers 100 languages. Note that extending the vocabulary may be necessary for new languages that are not included in the original tokenizer, while we do not extend the vocabulary in the experiments since all the languages in the experiments are included in the vocabulary of XLM-R, and we find that the unknown token rates for unseen languages in the experiments are zero.
Therefore, for all the models, we directly use mBART's tokenizer on the text for all languages in the experiments to ensure a fair comparison in BLEU, and we use thai-segmenter~\footnote{https://pypi.org/project/thai-segmenter/} to pre-tokenize the text in Thai (Th) before using mBART's tokenizer. For inference, we use beam search with a beam size of 5 for all directions.

\section{Results \& Analysis}
\subsection{Main Results}
% 1. mainly describe and analyze the results
The results of our proposed methods and baseline models are illustrated in Table~\ref{tab:main}, from which we can observe that conducting CPT on mBART is generally effective in the low-resource scenario of the NMT task, although the size of the pre-training corpus is as few as 100K paragraphs. Also, we can see that the CPT w/ MLT consistently outperforms all baseline models since the additional mixed-language information helps to construct a better alignment between the source and target languages in the translation pair. We observe that the CPT w/ MLT (Tgt) significantly outperforms mBART in multiple translation pairs (e.g., 2.92 BLEU points in En $\rightarrow$ Id and 2.39 BLEU points in En $\rightarrow$ Th).
We find that, although conducting CPT (w/ Ori or w/ MLT) on the text that contains tokens in the unseen language generally enhance the performance in the translation, the effectiveness of CPT w/ Ori is relatively deficient compared to CPT w/ MLT. We conjecture that the original objective function of mBART loses its advantages when the amount of pre-training monolingual data is small, while MLT is still beneficial thanks to the additional bilingual alignments that it have learned.
% Although mBART is shown to possess the ability to generalize to unseen languages~\cite{liu2020multilingual}, its performance will still be greatly limited, especially when both languages are unseen in the pre-training, due to the lack of language understanding ability on the unseen language. Conducting CPT on text that contains tokens in the unseen language can mitigate this issue and enhance the performance in the translation.
% generalization ability of the target translation pair.

% 2. pre-train direction (decode lang2 vs. decode lang1)
Additionally, we find that the direction of the CPT (Src or Tgt) also plays an important role. As we can see from Table~\ref{tab:main}, conducting CPT by reconstructing the target language in the translation pair generally achieves better performance than reconstructing the source. We conjecture that making the generated language in the CPT stage consistent with that in the fine-tuning stage will increase the benefits from the CPT. This is because, if the generated languages are different in these two stages, the model needs to learn to generate sentences on an entirely different language with only a few data samples in the fine-tuning stage, which could make the fine-tuning task much more challenging.
Interestingly, when English (a seen language) is the target language, the CPT w/ Ori (Tgt) becomes less effective, but CPT w/ MLT (Tgt) still works well. The reason is that CPT w/ Ori (Tgt) ignores the unseen language in the continual pre-training stage, while the mixed-language inputs of CPT w/ MLT (Tgt) still contain the tokens in the unseen language, which still enables the model to learn the unseen language.
% 3. mT5 vs. mBART
Surprisingly, mT5 performs generally worse than mBART, although it covers all the languages in our experiments. We conjecture that, since the objective function of mT5 is to generate the masked tokens, it makes the averaged length of the generated text relatively shorter than mBART, which might limit its ability to quickly adapt to a generation task in the low-resource scenario.

\begin{figure*}[!th]
\centering
\begin{subfigure}{.32\textwidth}
    \centering
    \includegraphics[scale=0.41]{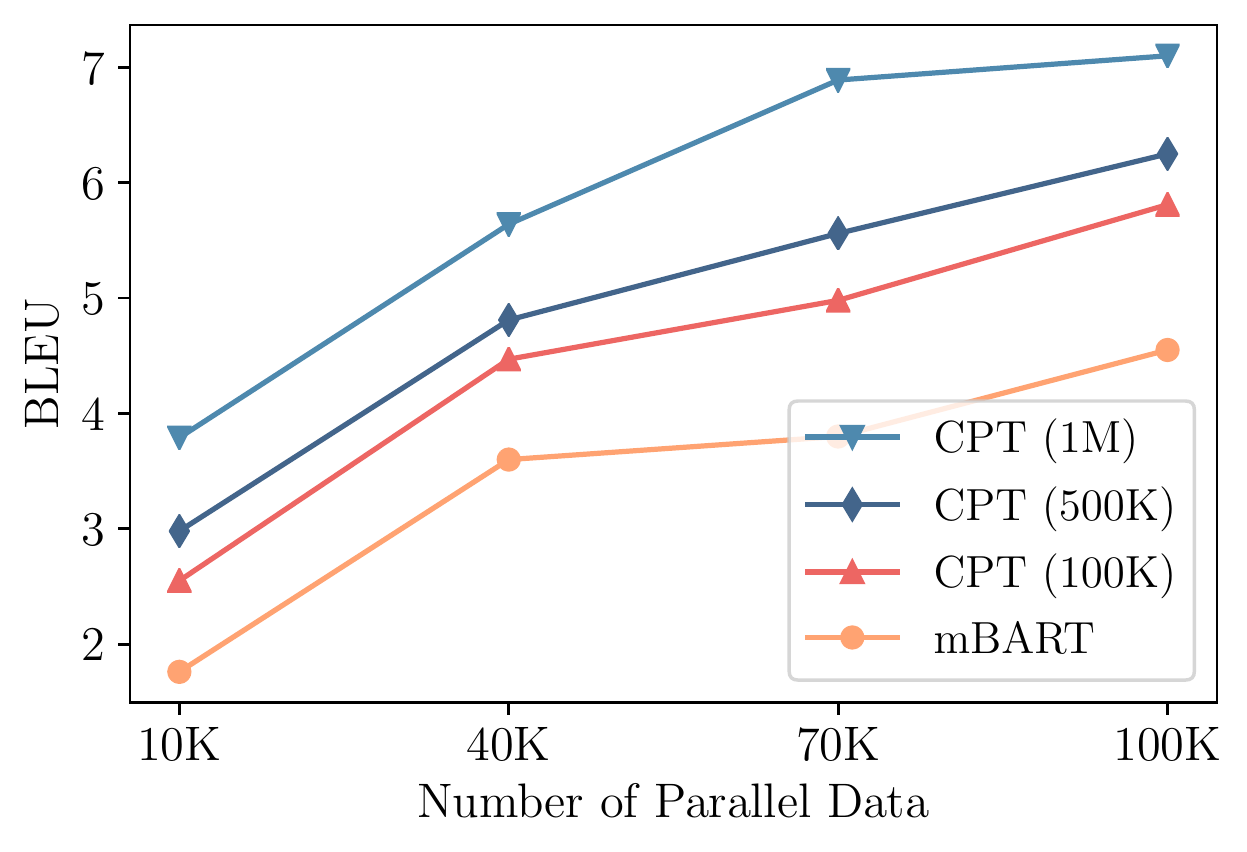}
    \caption{Sk $ \rightarrow $ Sv}
    \label{fig:sk_sv}
\end{subfigure}
\begin{subfigure}{.32\textwidth}
    \centering
    \includegraphics[scale=0.41]{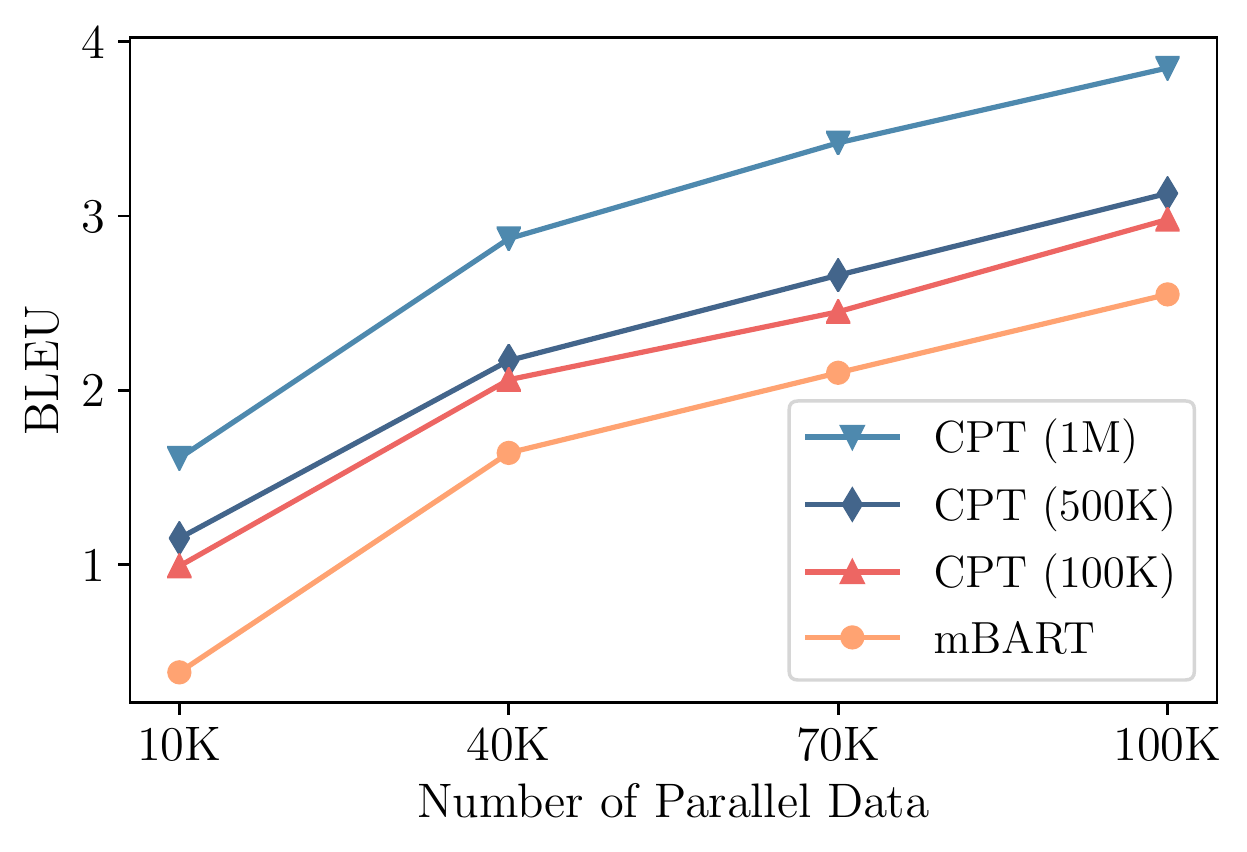}
    \caption{Sv $ \rightarrow $ Sk}
    \label{fig:sv_sk}
\end{subfigure}
\begin{subfigure}{.32\textwidth}
    \centering
    \includegraphics[scale=0.41]{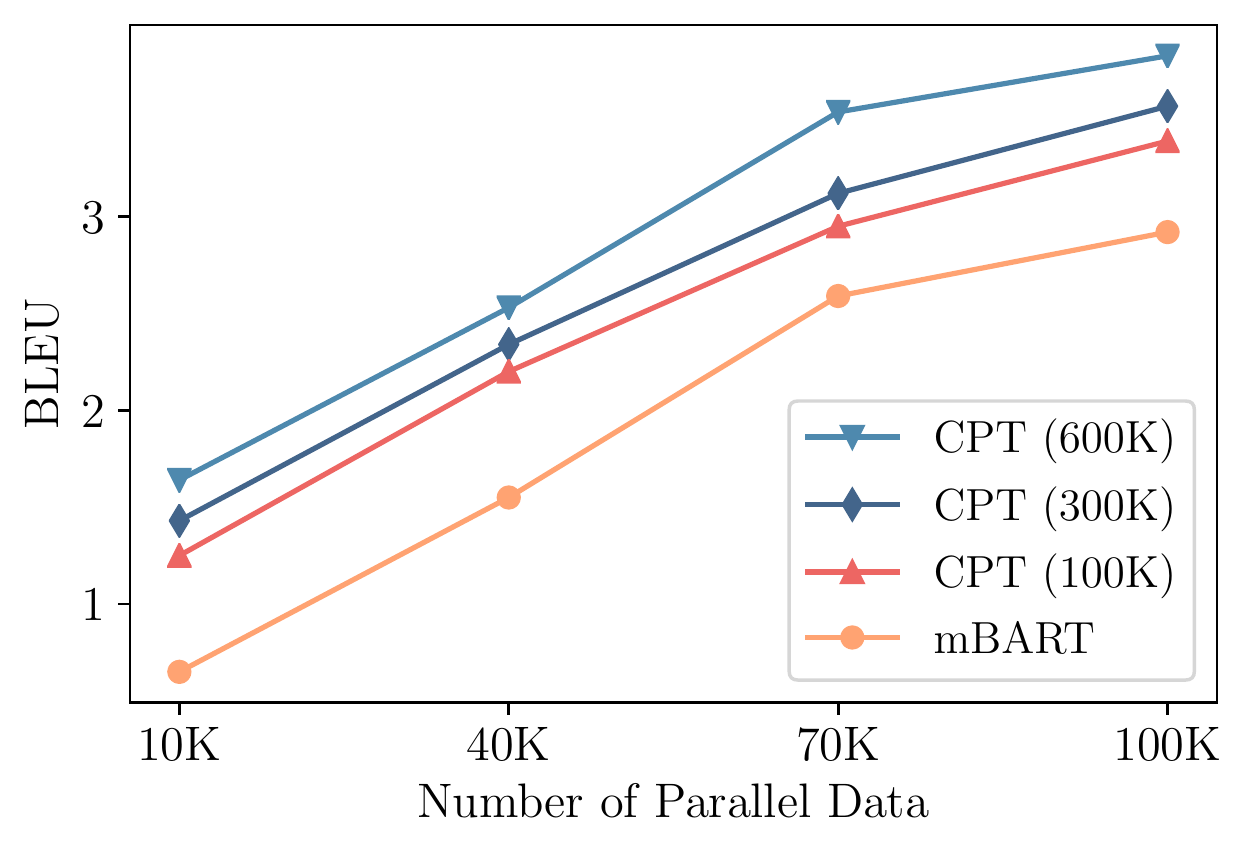}
    \caption{En $ \rightarrow $ Bn}
    \label{fig:en_bn}
\end{subfigure}
\begin{subfigure}{.32\textwidth}
    \centering
    \includegraphics[scale=0.41]{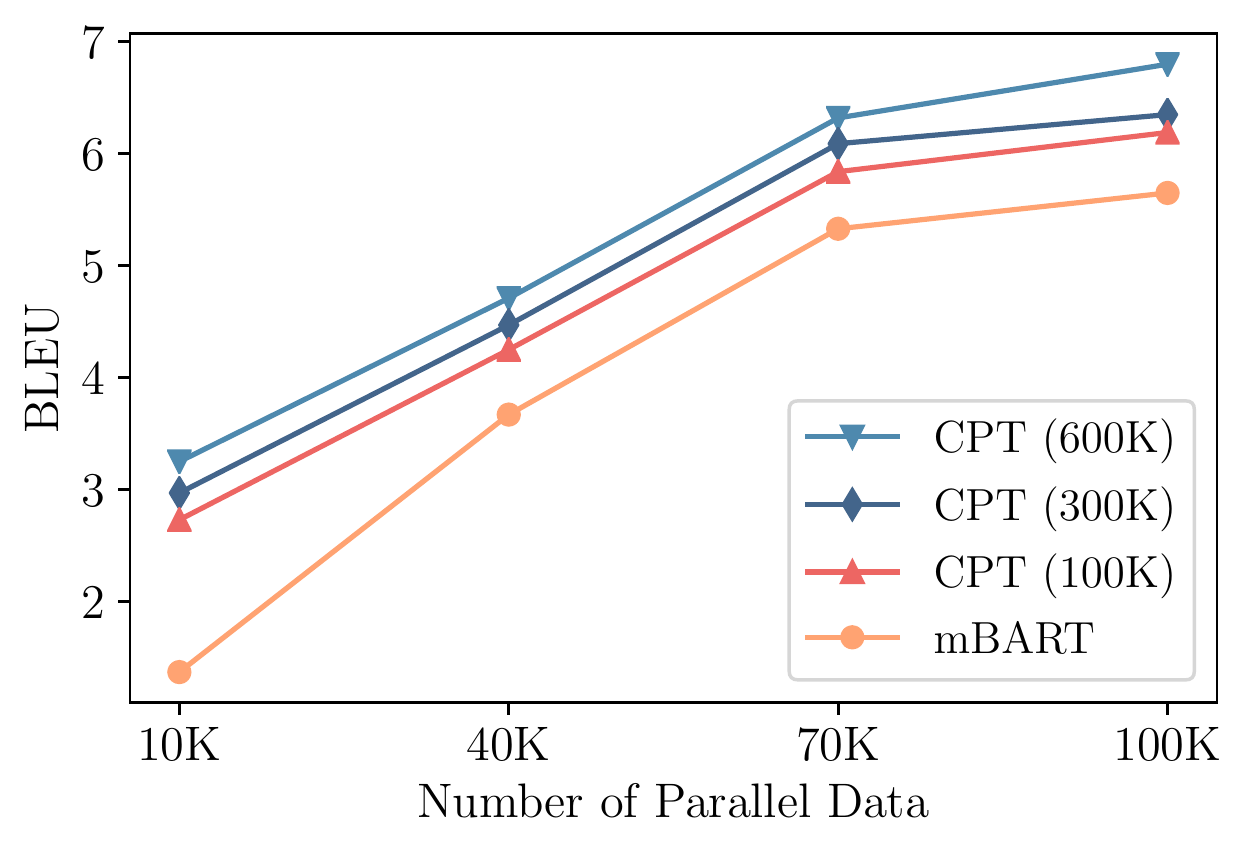}
    \caption{Bn $ \rightarrow $ En}
    \label{fig:bn_en}
\end{subfigure}
\begin{subfigure}{.32\textwidth}
    \centering
    \includegraphics[scale=0.41]{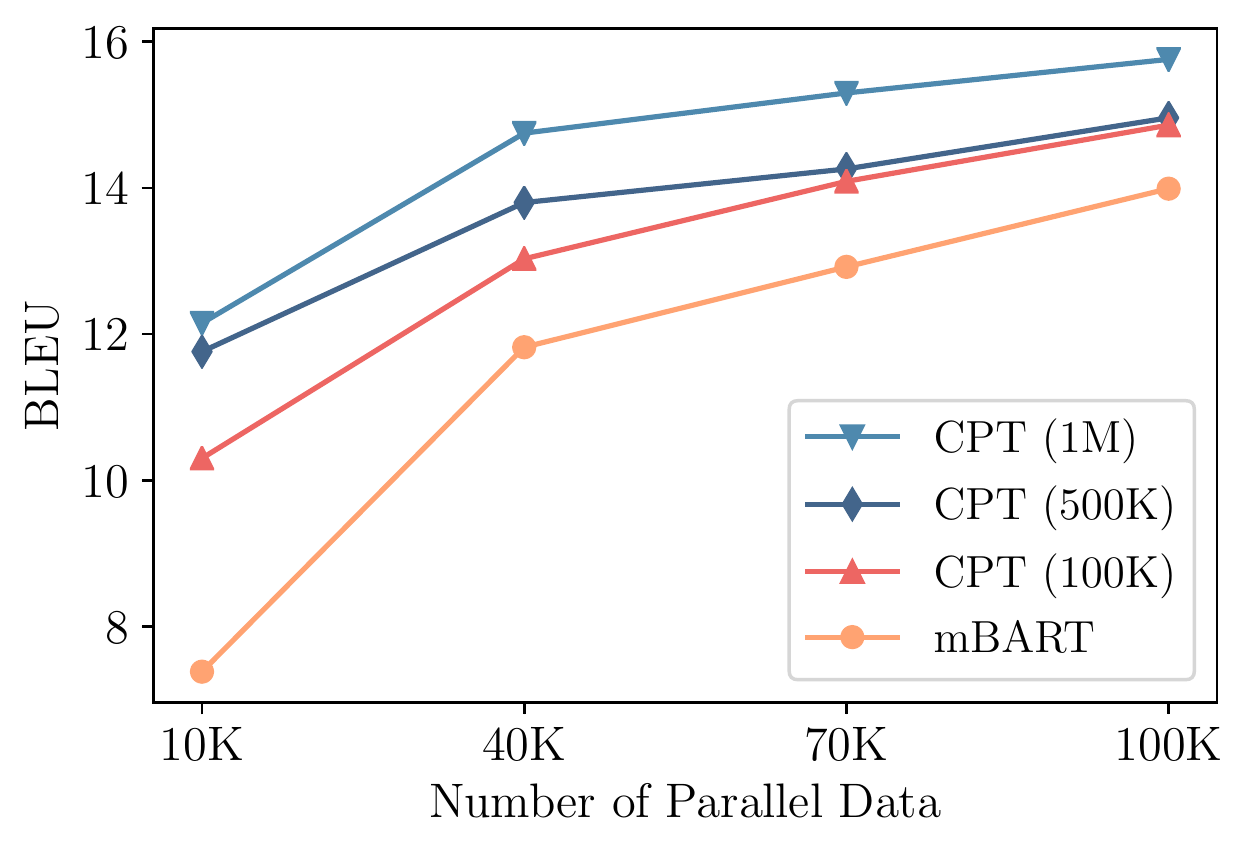}
    \caption{En $ \rightarrow $ Id}
    \label{fig:en_id}
\end{subfigure}
\begin{subfigure}{.32\textwidth}
    \centering
    \includegraphics[scale=0.41]{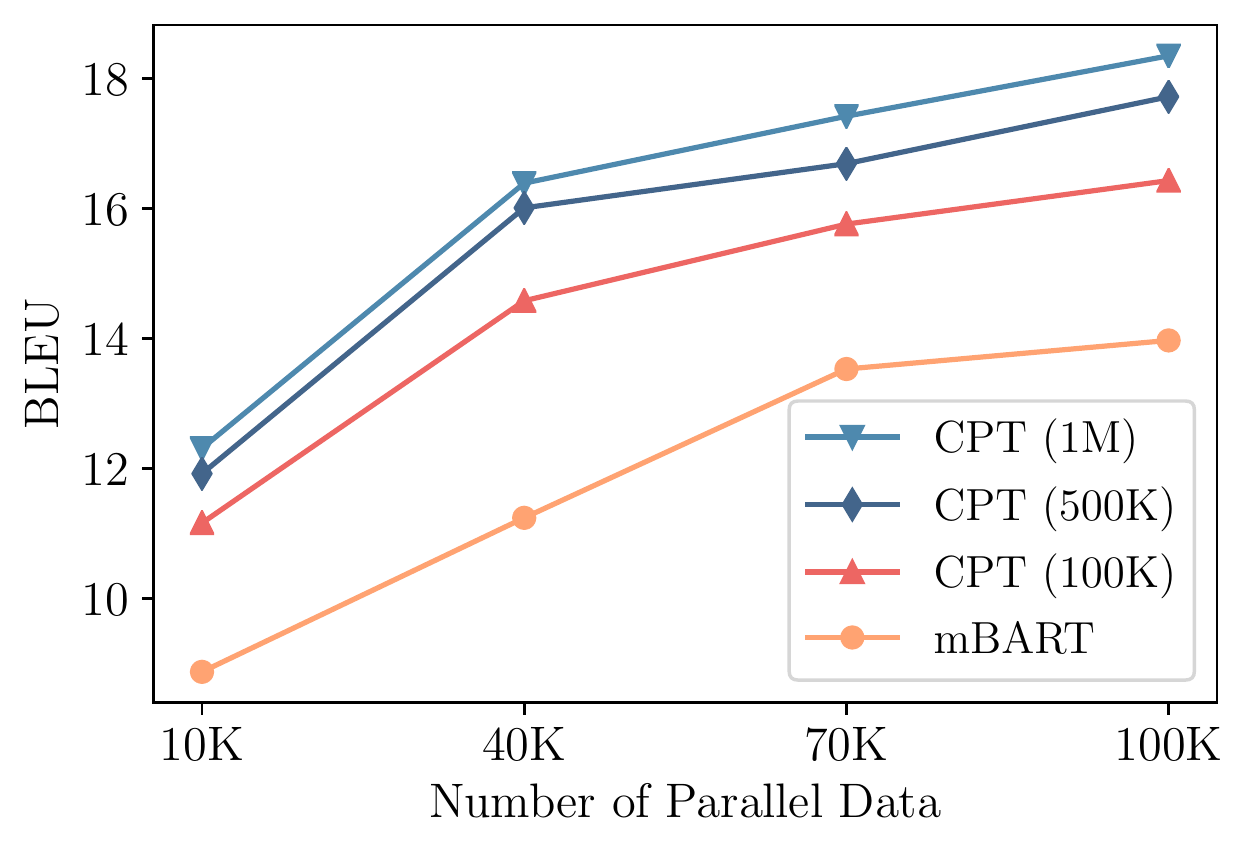}
    \caption{Id $ \rightarrow $ En}
    \label{fig:id_en}
\end{subfigure}
\caption{The performance over different numbers of parallel data (from 10K to 100K) and pre-training data (from 100K to 1M). CPT denotes our method, CPT w/ MLT (Tgt). Since the maximum number of paragraphs in Wikipedia for Bn is $\sim$600K, we set the data size in the CPT as 100K, 300K and 600K for En $\leftrightarrow$ Bn.}
\label{fig:diff_lowresource}
\end{figure*}

\subsection{Different Low-Resource Settings}
In this section, we investigate whether our method can generalize to other low-resource settings (i.e., different sizes of the parallel data and monolingual data). We choose three translation pairs (En $\leftrightarrow$ Bn, En $\leftrightarrow$ Id, and Sk $\leftrightarrow$ Sv), which cover two scenarios: 1) only one unseen language in a translation pair; and 2) both languages in a translation pair are unseen.
As illustrated in Figure~\ref{fig:diff_lowresource}, we can observe that our method is able to consistently improve on the mBART baseline in terms of different parallel data sizes, and the improvements can be further boosted when the size of the pre-training data (monolingual data) increases. This is because a larger corpus is able to amplify the benefits of MLT and better align the space between the two languages in the translation.
Moreover, we find that our method is especially effective for the Sk $\leftrightarrow$ Sv translation pair when the size of the pre-training data reaches 1M. For example, in the Sk $\rightarrow$ Sv translation, the performance of CPT (1M) with 10K parallel samples (3.79) is on par with mBART with 70K parallel samples (3.80), which might suggest that gathering larger monolingual data (along with a dictionary) can be an alternative to collecting a larger size of parallel data.

\begin{table}[!th]
\renewcommand{\arraystretch}{1.15}
\centering
\begin{adjustbox}{width={0.49\textwidth},totalheight={\textheight},keepaspectratio}
\begin{tabular}{c|cccc}
\Xhline{2\arrayrulewidth}
\textbf{Language Pairs}  & \multicolumn{2}{c}{\textbf{En-Gu}} & \multicolumn{2}{c}{\textbf{En-Kk}} \\
\textbf{Direction}   & $\leftarrow$  & $\rightarrow$ & $\leftarrow$ & $\rightarrow$        \\ \hline
mBART       & 3.11        & 0.10        & 8.93         & 2.44       \\ \hline
CPT w/ MLT (Tgt, 100K) & 4.59        & 2.01        & 10.16        & 4.01       \\
CPT w/ MLT (Tgt, 500K) & 5.44        & 2.91        & 10.74         & 4.74       \\
CPT w/ MLT (Tgt, 1M)   & \textbf{5.97}        & \textbf{3.89}        & \textbf{11.45}         & \textbf{5.29}       \\ \Xhline{2\arrayrulewidth}
\end{tabular}
\end{adjustbox}
\caption{The effectiveness of our method on seen languages. 100K, 500K and 1M are the corpus sizes for the CPT w/ MLT (Tgt).}
\label{tab:seen}
\end{table}

\begin{table}[!th]
\renewcommand{\arraystretch}{1.15}
\centering
\begin{adjustbox}{width={0.45\textwidth},totalheight={\textheight},keepaspectratio}
\begin{tabular}{c|ccc}
\Xhline{2\arrayrulewidth}
 \textbf{Models}   & \textbf{mBART} & \textbf{CPT w/ Ori} & \textbf{CPT w/ MLT} \\ \hline
Avg & 0.00  & 0.27        & \textbf{0.53}         \\ \Xhline{2\arrayrulewidth}
\end{tabular}
\end{adjustbox}
\caption{Averaged performance over the 24 translation pairs in the zero-shot test. Both CPT methods are to reconstruct the target language with 100K samples.}
\label{tab:zeroshot}
\end{table}

\subsection{Effectiveness on Seen Languages}
As we can see from Table~\ref{tab:seen}, the CPT w/ MLT can also significantly improve the performance on the translation pairs where both languages are in the mBART's pre-training list. The CPT w/ MLT improves by at least 1.2 BLEU points on all translation pairs with only 100K pre-training data. Additionally, the improvement brought by our method can be further boosted when a larger pre-training corpus is available, which accords with the experimental results for the unseen languages.

We conjecture two reasons: 1) Continuing pre-training mBART can make the model focus on the languages in the translation pair and increase the model's ability of fast adaptation to the translation task. 2) Continual pre-training with the mixed-language text can further align the two languages in the translation, which gives a better initialization for the low-resource translation task.

\begin{table*}[!th]
\renewcommand{\arraystretch}{1.15}
\centering
\begin{adjustbox}{width={0.82\textwidth},totalheight={\textheight},keepaspectratio}
\begin{tabular}{l|cccccccc}
\Xhline{2\arrayrulewidth}
\multicolumn{1}{c|}{\textbf{Language Pairs}} & \multicolumn{2}{c}{\textbf{En-Id}} & \multicolumn{2}{c}{\textbf{En-Af}} & \multicolumn{2}{c}{\textbf{Id-Ta}} & \multicolumn{2}{c}{\textbf{Sk-Sv}} \\
\multicolumn{1}{c|}{\textbf{Direction}}      & $\leftarrow$  & $\rightarrow$ & $\leftarrow$ & $\rightarrow$   & $\leftarrow$  & $\rightarrow$ & $\leftarrow$ & $\rightarrow$      \\ \hline
\multicolumn{1}{c|}{mBART}          & 8.87        & 7.38        & 8.24        & 10.02       & 1.21        & 0.98        & 0.38        & 1.76        \\ \hline
CPT w/ MLT (Tgt)                   & \textbf{11.16}       & \textbf{10.30}       & \textbf{10.56}       & \textbf{11.62}       & \textbf{2.52}        & \textbf{1.75}        & \textbf{0.99}        & \textbf{2.55}        \\
\hspace{2mm} w/o noise                           & 11.06       & 9.94        & 8.95        & 10.65       & 2.11        & 1.29        & 0.79        & 1.95        \\
\hspace{2mm} w/o deletion                    & 10.33       & 9.57        & 8.62        & 10.24       & 1.90        & 1.19        & 0.86        & 2.13        \\
\hspace{2mm} w/o noise \& deletion                 &  10.12   & 9.03   & 8.79   & 10.08  & 1.84 & 1.07    & 0.73 & 1.92   \\ \Xhline{2\arrayrulewidth}
\end{tabular}
\end{adjustbox}
\caption{Ablation study on the noise function $g$ (denoted as noise) and token deletion (denoted as deletion).}
\label{tab:noise_deletion}
\end{table*}

\begin{table}[!th]
\renewcommand{\arraystretch}{1.15}
\centering
\begin{adjustbox}{width={0.49\textwidth},totalheight={\textheight},keepaspectratio}
\begin{tabular}{c|cccc}
\Xhline{2\arrayrulewidth}
\textbf{Language Pairs}     & \multicolumn{2}{c}{\textbf{En-Id}} & \multicolumn{2}{c}{\textbf{En-Th}} \\
\textbf{Direction}  & $\leftarrow$  & $\rightarrow$ & $\leftarrow$ & $\rightarrow$  \\ \hline
mBART              & 8.87        & 7.38        & 3.12        & 2.41        \\ \hline
CPT w/ MLT (10\%) & 9.65    & 9.14        & 3.28            & 3.15            \\
CPT w/ MLT (20\%) & 9.56    & 9.98        & 3.49            & 3.78            \\
CPT w/ MLT (30\%) & \textbf{11.16}       & 10.30       & \textbf{3.85}        & \textbf{4.54}        \\
CPT w/ MLT (40\%) & 10.78    & \textbf{10.34}       & 3.56            & 4.49   \\
CPT w/ MLT (50\%) & 10.06   & 9.97        & 3.15     & 3.67  \\ \Xhline{2\arrayrulewidth}
\end{tabular}
\end{adjustbox}
\caption{Effectiveness of CPT w/ MLT (Tgt) in terms of different language mixing ratios. The ratio in the brackets denotes the number of source language tokens (in the translation pair) divided by that of the target language tokens.}
\label{tab:ratio}
\end{table}

\begin{table}[!th]
\renewcommand{\arraystretch}{1.15}
\centering
\begin{adjustbox}{width={0.46\textwidth},totalheight={\textheight},keepaspectratio}
\begin{tabular}{c|cccc}
\Xhline{2\arrayrulewidth}
\textbf{Language Pairs}   & \multicolumn{4}{c}{\textbf{En-Th}}                               \\ \hline
\textbf{Direction}        & \multicolumn{2}{c|}{$\longleftarrow$}            & \multicolumn{2}{c}{$\longrightarrow$} \\
\textbf{Pre-Tokenization}   & \cmark & \multicolumn{1}{c|}{\xmark}     & \cmark          & \xmark         \\ \hline
mBART            & \textbf{3.12} & \multicolumn{1}{c|}{3.04} & \textbf{2.41}      & 0.43     \\ \hline
CPT w/ Ori (Src) & 3.17 & \multicolumn{1}{c|}{\textbf{3.18}} & \textbf{3.16}      & 0.53     \\
CPT w/ Ori (Tgt) & 3.08 & \multicolumn{1}{c|}{\textbf{3.09}} & \textbf{3.57}      & 0.56     \\ \hline
CPT w/ MLT (Src) & \textbf{3.42} & \multicolumn{1}{c|}{3.37} & \textbf{4.80}      & 0.64     \\
CPT w/ MLT (Tgt) & \textbf{3.85} & \multicolumn{1}{c|}{3.79} & \textbf{4.54}      & 0.70     \\ \Xhline{2\arrayrulewidth}
\end{tabular}
\end{adjustbox}
\caption{Comparison between conducting and not conducting the pre-tokenization for Thai.}
\label{tab:pretokenization}
\end{table}

\subsection{Zero-shot Performance}
To further analyze the alignment quality between the source and target languages in the translation after the CPT, we evaluate the models in the zero-shot scenario, where we directly test the pre-trained models on the test set without any fine-tuning on the parallel data. As illustrated in Table~\ref{tab:zeroshot}, we can see that the zero-shot performance is relatively low since the models are not trained on any parallel or pseudo-parallel data~\footnote{The results for each translation pair are in Appendix~\ref{sec:appendix_a}.}, and mBART gets 0 BLEU points due to the unseen languages in the test data. We find that CPT w/ Ori achieves more than 0 BLEU points, even though it does not utilize any supervision from the bilingual text. We conjecture that this can be attributed to the multilingual ability of mBART. Furthermore, CPT w/ MLT is able to outperform CPT w/ Ori since it learns additional bilingual alignments by reconstructing the target documents from the mixed-language text. In addition, the results are able to further illustrate that our method is able to achieve a better alignment quality than the baseline method.

\subsection{Ablation Study}
In this section, we first explore how the noise function $g$ and token deletion in function $h$ affect the effectiveness of our method ($g$ and $h$ are described in~§\ref{noisy_cs_function}). Then, we investigate how the language mixing ratio of the mixed-language text affects our method's performance.

\paragraph{Noise \& Deletion}
As shown in Table~\ref{tab:noise_deletion}, we can see that both the noise function and token deletion play an important part in the CPT, and removing both of them further degrades the performance. 
Given that the number of pre-training documents is as few as 100K, it is relatively difficult for the model to learn a good representation for the unseen language.
However, adding the noise function in the CPT forces the model to learn to perform text infilling and sentence reordering, which increases the model's ability to understanding the unseen language. 
Conducting the token deletion brings two benefits: 1) It increases the variety of the mixed-language text, which makes the model not overfit to a certain mixed-language pattern. 2) It also injects extra noise to the inputs, which further compels mBART to understand the unseen language better.
Moreover, incorporating both noise function $g$ and the token deletion further boosts the effectiveness of the pre-training.

\paragraph{Language Mixing Ratio}
We control the probability of whether to replace a token with its translation to generate different settings of language mixing ratios and investigate how different ratios affect the effectiveness of the pre-training.
As shown in Table~\ref{tab:ratio}, using a too high or too low mixing ratio will degrade the advantages of the CPT w/ MLT,~\footnote{Note that the maximum mixing ratio will not be larger than 55\% since there are substantial infrequent tokens in the target language not existing in the dictionary, which will not be replaced with the source language tokens.} and keeping the ratio between 30\% and 40\% will achieve the best performance. 
We conjecture that, if the mixing ratio is too low, the dictionary which provides the supervision of bilingual alignment is not well utilized, while if the mixing ratio is too high (e.g., 50\%), we replace almost all the tokens existing in the dictionary, which lowers the diversity of the mixed-language text and makes the model more easily overfit to the pre-training data.

\subsection{Importance of Pre-Tokenization}
Considering that the tokenizer of mBART is created based on the text of the pre-training languages, it might not perform good tokenization for the unseen languages that are diverse from the pre-trained languages. Therefore, it could be a better option to pre-tokenize the text before using mBART's tokenizer. We conduct experiments on the En-Th language pair and compare the performance between performing and not performing the pre-tokenization for Thai. As shown in Table~\ref{tab:pretokenization}, we find that pre-tokenization is able to improve the performance in En $\rightarrow$ Th significantly, while the improvements are marginal in Th $\rightarrow$ En. We conjecture that decoding (generating) tokens in the unseen language is much more difficult than encoding those tokens when they are not properly tokenized. This is because the task of the encoder is to understand the meaning of the input text, while the decoder needs to attend to the input text and generate tokens simultaneously, which makes the task of the decoder more difficult than that of the encoder.
Therefore, when the unseen language (Thai) becomes the target language in the translation pair, the performance drops remarkably without pre-tokenization.

\section{Related Work}
\subsection{Multilingual Pre-Trained Models}
Recently, multilingual pre-trained models based on the masked language modeling (MLM) objective function~\cite{devlin2019bert,conneau2019cross,huang2019unicoder,conneau2020unsupervised} have shown their effectiveness at performing cross-lingual classification-based tasks. However, these models are inferior to the generation tasks~\cite{ronnqvist2019multilingual} since they are not pre-trained in a generative way. 
Multilingual pre-training performed in a Seq2Seq fashion is able to mitigate this issue~\cite{radford2019language,lewis2020bart,raffel2019exploring}, and has become a strong backbone for building NMT systems, especially in a low-resource scenario~\cite{liu2020multilingual,song2019mass,lin2020pre,yang2020csp,xue2020mt5,fan2020beyond,tang2020multilingual}.
\citet{liu2020multilingual} pre-trained a Seq2Seq multilingual model (mBART) by denoising full texts in 25 languages, while \citet{lin2020pre} proposed multilingual random aligned substitution to pre-train an NMT model for many languages based on parallel data.
Instead of pre-training models from scratch, \citet{wang2020extending} proposed to extend multilingual BERT~\cite{devlin2019bert} to an unseen language and evaluate it on the named entity recognition task.
Although many studies have focused on pre-training multilingual models, few have investigated how to adapt the pre-trained models to new languages effectively. Also, to the best of our knowledge, we are the first to explore how to adapt a multilingual model pre-trained in a Seq2Seq fashion to unseen languages and evaluate the methods on a generative task (the NMT task).

\subsection{Low-Resource Machine Translation}
Recently, developing algorithms that are able to cope with the scenario where the training data are insufficient have become an interesting and popular research topic across a variety of tasks~\cite{chen2019meta,chen2019closer,chen2020few,brown2020language,liu2020attention,lauscher2020zero,winata2020learning,liu2020importance,peng2020few,liu2020crossner,yu2021adaptsum,winata2021multilingual}.
Low-resource machine translation systems~\cite{vandeghinste2007metis,irvine2013combining,zoph2016transfer,sennrich2016improving,fadaee2017data,currey2017copied,imankulova2017improving,gu2018universal,pourdamghani2018using,gu2018meta,lample2018unsupervised,lample2018phrase,kocmi2018trivial,artetxe2018unsupervised,lakew2018multilingual,imankulova2019exploiting,xia2019generalized,liu2019incorporating,guzman2019flores,imankulova2019filtered,stickland2020recipes,siddhant2020leveraging} alleviated the parallel data scarcity issue for low-resource languages and improve the models' generalization ability for low-resource language pairs. \citet{pourdamghani2018using} proposed to improve the low-resource NMT performance by boosting the quality of word alignments. \citet{gu2018meta} applied the meta-learning approach into the low-resource NMT task, and \citet{baziotis2020language} incorporated a language model prior to regularize the output distribution of the translation model. Pre-training a multilingual Seq2Seq model~\cite{liu2020multilingual,lin2020pre} allows it to be directly fine-tuned for supervised machine translation tasks and produces remarkable performance gains in the low-resource scenario over those without pre-training.

\section{Conclusion \& Future Work}
In this paper, we present a continual pre-training framework to improve mBART's generalization ability to extremely low-resource translation pairs that contain unseen languages. 
We propose to construct noisy mixed-language text from the monolingual corpus to cover both the source and target languages, and then, we continue pre-training mBART to reconstruct the original monolingual text. Results illustrate that our method is able to consistently surpass strong baselines across all tested translation pairs that contain unseen languages, as well as the ones where both languages are seen in the original mBART's pre-training.
Moreover, we observe that our method is also beneficial for different low-resource settings, and its performance can be further boosted when a larger pre-training corpus is available. Furthermore, we find that not only mixing the source and target languages, but also increasing the variety of the inputs plays an essential role in the continual mixed-language pre-training.
In future work, we will explore more pre-training methods to further boost the performance of pre-trained models on the NMT task. Additionally, we will study more applications of continual mixed-language pre-training, such as applying it to downstream cross-lingual tasks.

\section*{Acknowledgement}
We want to say thanks to the anonymous reviewers for the insightful reviews and constructive feedback. This work is partially funded by ITF/319/16FP and MRP/055/18 of the Innovation Technology Commission, the Hong Kong SAR Government.

\bibliographystyle{acl_natbib}
\bibliography{acl2021}

\clearpage
\appendix

\section{Zero-shot Performance}
\label{sec:appendix_a}
The zero-shot performance for the 24 translation pairs are shown in Table~\ref{tab:zeroshot_full} (in the next page). We find that the CPT w/ MLT generally outperforms the CPT w/ Ori. However, the zero-shot results are relatively low, especially for the translation pairs where both languages are unseen in the original mBART's pre-training, due to the absence of parallel data.

\begin{table*}[t!]
\renewcommand{\arraystretch}{1.19}
\centering
\begin{adjustbox}{width={0.999\textwidth},totalheight={\textheight},keepaspectratio}
\begin{tabular}{c|cccccccccccc}
\Xhline{2\arrayrulewidth}
\textbf{Language Pairs}          & \multicolumn{2}{c}{\textbf{En-Id}} & \multicolumn{2}{c}{\textbf{En-Uk}} & \multicolumn{2}{c}{\textbf{En-Bn}} & \multicolumn{2}{c}{\textbf{En-Af}} & \multicolumn{2}{c}{\textbf{En-Ta}} & \multicolumn{2}{c}{\textbf{En-Th}} \\
\textbf{Direction}   & $\leftarrow$  & $\rightarrow$ & $\leftarrow$ & $\rightarrow$  & $\leftarrow$  & $\rightarrow$ & $\leftarrow$ & $\rightarrow$ & $\leftarrow$  & $\rightarrow$ & $\leftarrow$ & $\rightarrow$       \\ \hline
mBART                   & 0.04        & 0.00        & 0.02        & 0.00        & 0.00        & 0.00        & 0.00        & 0.00        & 0.00        & 0.00        & 0.00        & 0.00        \\
CPT w/ Ori (Tgt)        & 1.35        & 1.16        & 0.46        & \textbf{0.43}        & 0.07        & \textbf{0.39}        & 0.95        & 0.78        & 0.06        & 0.13        & 0.16        & 0.38        \\
CPT w/ MLT (Tgt)        & \textbf{1.80}        & \textbf{1.24}        & \textbf{1.68}        & 0.41        & \textbf{0.56}        & 0.36        & \textbf{2.43}        & \textbf{1.52}        & \textbf{0.54}        & \textbf{0.17}        & \textbf{1.02}        & \textbf{0.52}        \\ \hline \hline
\textbf{Language Pairs} & \multicolumn{2}{c}{\textbf{Id-Ta}} & \multicolumn{2}{c}{\textbf{Bn-Th}} & \multicolumn{2}{c}{\textbf{Bg-Ta}} & \multicolumn{2}{c}{\textbf{Id-Bn}} & \multicolumn{2}{c}{\textbf{Mk-Th}} & \multicolumn{2}{c}{\textbf{Sk-Sv}} \\
\textbf{Direction}   & $\leftarrow$  & $\rightarrow$ & $\leftarrow$ & $\rightarrow$ & $\leftarrow$  & $\rightarrow$ & $\leftarrow$ & $\rightarrow$ & $\leftarrow$  & $\rightarrow$ & $\leftarrow$ & $\rightarrow$       \\ \hline
mBART                   & 0.00        & 0.00        & 0.00        & 0.00        & 0.00        & 0.00        & 0.00        & 0.00        & 0.00        & 0.00        & 0.00        & 0.00        \\
CPT w/ Ori (Tgt)        & 0.00        & 0.00        & \textbf{0.02}        & 0.00        & 0.00        & 0.00        & 0.06        & 0.04        & 0.00        & 0.02        & 0.00        & 0.00        \\
CPT w/ MLT (Tgt)        & 0.00        & 0.00        & 0.00        & 0.00        & 0.00        & 0.00        & \textbf{0.14}        & \textbf{0.05}        & \textbf{0.07}        & \textbf{0.03}        & 0.00        & \textbf{0.20}        \\ \Xhline{2\arrayrulewidth}
\end{tabular}
\end{adjustbox}
\caption{Zero-shot results for the 24 translation pairs.}
\label{tab:zeroshot_full}
\end{table*}

\section{Data \& Code}
We will release our split data, dictionaries, as well as the code to ensure the reproducibility of our work.

\end{document}